\documentclass[]{bytedance_seed}

\usepackage[toc,page,header]{appendix}
\usepackage{amssymb}
\usepackage{siunitx}

\title{VoT: Vision-of-Thought for Unified Multimodal Representation Alignment}
\author[*]{Jingxiang Sun}
\author[*\dagger]{Chao Liao}
\author[*]{Zhengxiong Luo}
\author{Chaorui Deng}
\author{Chen-lin Zhang}
\author{Junke Wang}
\author{Ceyuan Yang}
\author{Haoqi Fan}
\author[\ddagger]{Weilin Huang} 

\affiliation[]{ByteDance Seed}
\contribution[*]{Equal Contribution}
\contribution[\dagger]{Project Lead}
\contribution[\ddagger]{Corresponding Author}

\abstract{
Current text-to-image systems typically employ a ``text encoder plus diffusion decoder'' paradigm, in which text semantics directly modulate continuous latent noise. Despite their success, these methods lack an explicit, interpretable intermediate representation that effectively bridges high-level linguistic semantics and low-level visual signals. In this paper, we propose Vision-of-Thought (VoT), a framework that introduces a discrete visual-thinking layer between vision-language models (VLMs) and diffusion transformers (DiTs). Instead of treating VLMs merely as text encoders, we use them as multimodal planners that generate discrete VoT tokens representing high-level visual plans, such as objects and layouts, before rendering pixels. We train a specialized VoT tokenizer in the VLM semantic space with a closed-loop objective that combines VLM alignment, feature reconstruction, and vector-quantization losses. These objectives make the tokens semantically readable by the VLM while preserving the visual information needed for generation. Experimental results demonstrate that VoT improves semantic alignment and provides a structured interface for interpretable and controllable generation.}

\date{\today}

\begin{document}
\maketitle

\section{Introduction}

% 当前主流的文生图模型多采用：
% - Text Encoder + DiT 两阶段范式：文本 → text embedding/KV → 直接调控扩散过程。
% - 引入 VLM 之后，VLM 通常仍被当作“更强的文本编码器/统一编码器”，文本语义直接作用在连续噪声/latent 上。

% 这类范式在 文本 token 与图像 latent 之间缺乏一个显式、可组合、可编辑的 **视觉中间语**，导致：
% - 跨模态bridge主要依赖隐式 cross-attention；
% - 生成过程可解释性、可控性有限；
% - 难以在统一的离散空间中做多模态推理与生成

\begin{figure*}[t!]
    \centering
    \includegraphics[width=1.0\linewidth]{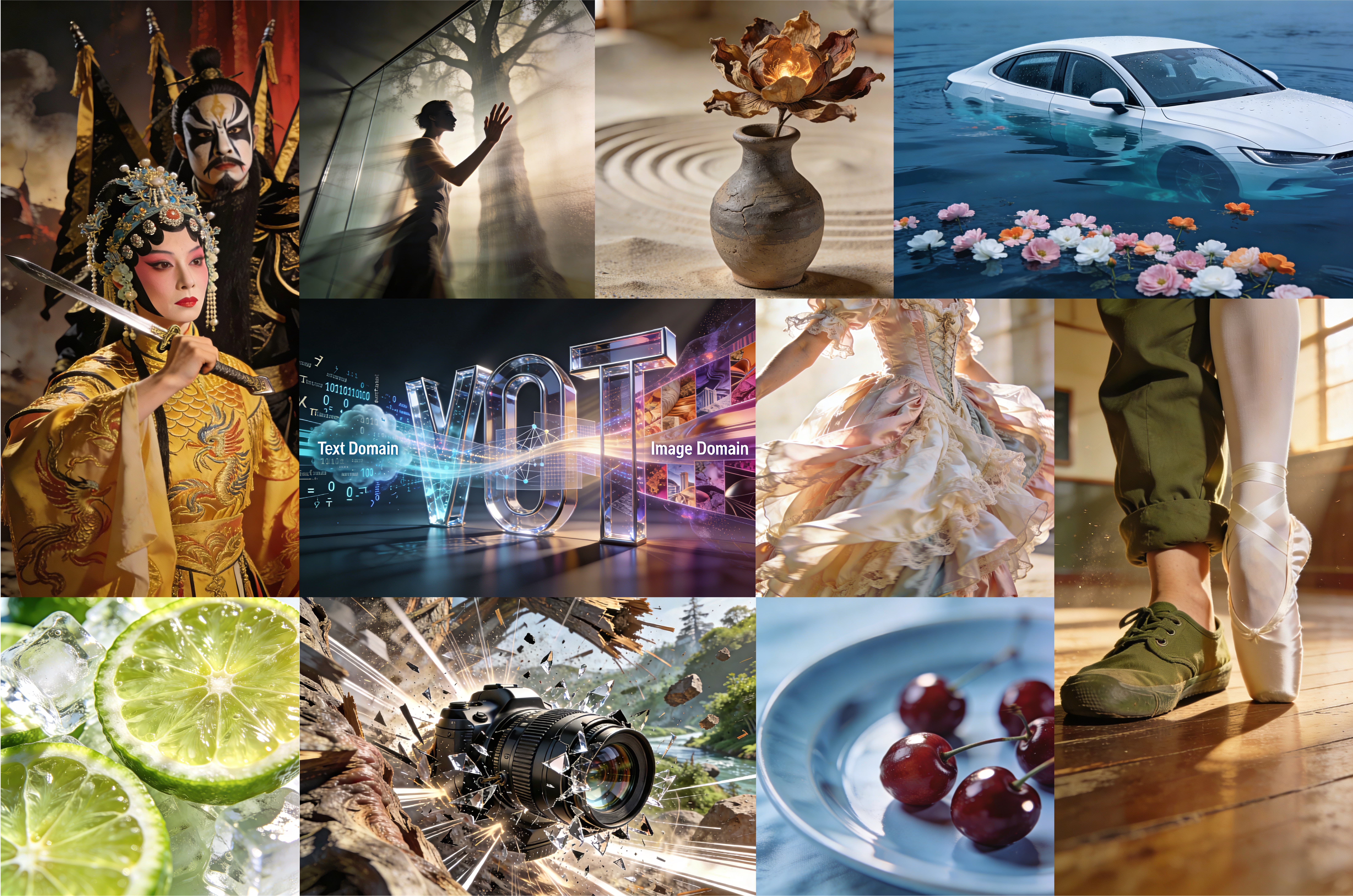}
    \caption{Images generated by VoT from complex text and image prompts.}
    \label{fig:teaser}
\end{figure*}

A unified model capable of both deep understanding and high-fidelity generation holds immense potential for the future of AI. Such a system could seamlessly generate images based on complex instructions, reason about visual data, and visualize multimodal analyses through generated outputs. The unveiling of GPT-4o's enhanced capabilities has further highlighted this potential, sparking widespread interest in unification.

However, designing such a unified framework presents a formidable challenge. It requires integrating the specific strengths of autoregressive models---which excel at discrete reasoning and text generation---with the robustness of diffusion-based models for high-quality image synthesis. Current text-to-image (T2I) paradigms typically treat this as a direct mapping problem, encoding text prompts into static embeddings or key--value (KV) caches that condition a diffusion process. While effective for simple descriptions, this approach faces a significant modality gap. The rich, structured world knowledge embedded in a large language model (LLM) is difficult to transfer effectively to the diffusion decoder through static embeddings alone, leading to a bottleneck where the model ``knows'' what to generate but fails to ``render'' it accurately.

In this work, we introduce Vision-of-Thought (VoT) as a semantic bridge to align these distinct modalities. Adopting a ``parallel expert'' philosophy, we freeze a powerful vision-language model (VLM) to preserve its multimodal understanding capabilities and introduce lightweight autoregressive experts to generate discrete semantic visual tokens. Unlike recent query-based approaches (e.g., MetaQuery~\cite{pan2025transfer}, BLIP-3o~\cite{li2024blip3}) that generate parallel queries for conditioning, VoT generates visual tokens autoregressively. This choice is deliberate: while query-based methods offer parallel efficiency, they lack the structural consistency with VLMs required for complex reasoning. By treating visual generation as a next-token prediction task, VoT allows the model to perform ``visual chain-of-thought,'' planning the image composition step by step before rendering---effectively extending the VLM's reasoning capability into the visual domain.

A critical hurdle for autoregressive (AR) image generation is visual tokenization. Standard visual tokenizers (e.g., VQ-VAE, VQGAN) are optimized primarily for reconstruction fidelity, compressing images into ``pixel codes'' dominated by local textures and low-level details. As a result, their codebooks are fundamentally misaligned with the semantic representation space of VLMs: token sequences are often excessively long and semantically redundant, behaving more like compressed pixel fragments than meaningful semantic units. Moreover, these reconstruction-centric tokens tend to exhibit unfavorable statistics for next-token prediction (e.g., heavy-tailed usage and weak semantic correlation among neighbors), making AR modeling unstable. Consequently, when a VLM is tasked with predicting such tokens, it often struggles to maintain semantic coherence, which degrades downstream generation quality.

To address this mismatch, we propose a \emph{VoT Tokenizer} that is explicitly aligned with the VLM semantic space through a \emph{closed-loop alignment} paradigm. Instead of quantizing raw pixels, we perform \emph{semantic quantization} by using the frozen ViT features of a pretrained VLM as the quantization target, ensuring discrete tokens originate from a high-level semantic manifold. We optimize the tokenizer with three objectives: (i) a \emph{VLM alignment loss} that enables the VLM to ``read'' and use the quantized tokens, (ii) a \emph{feature reconstruction loss} that recovers the teacher ViT feature space, and (iii) a \emph{vector-quantization loss} that stabilizes codebook learning. The first two objectives are strongly complementary: VLM alignment provides language-grounded supervision, while reconstruction supplies dense, information-preserving constraints. Furthermore, we observe that \emph{scaling the teacher VLM} consistently improves tokenizer quality and downstream semantic alignment, suggesting that stronger teachers provide cleaner and more compositional supervision.

On the model side, we introduce an efficient architecture that integrates VoT into modern unified models by extending \emph{Mixture-of-Transformers (MoT)}. Specifically, we initialize a VoT branch by copying a subset of pretrained VLM experts and keep the original VLM branch frozen throughout training. This design preserves the VLM's multimodal understanding capability while enabling efficient visual-planning training with minimal additional compute. After tokenizer training, our model training proceeds in two stages: (i) \emph{VoT branch pretraining}, where the model learns to produce structured VoT tokens from text--image pairs, and (ii) \emph{VoT-DiT joint training}, where VoT prediction and diffusion-based rendering are jointly optimized so that the tokens serve as explicit visual plans.

Our contributions are summarized as follows:
% \begin{itemize}
%     \item Vision-of-Thought Framework: We propose VoT, a unified framework that decouples visual reasoning from rendering. By utilizing autoregressive experts for visual planning, we enable the transfer of complex reasoning capabilities from frozen VLMs to image generation.

%     \item VLM-Aligned Tokenizer: We introduce a semantic-aligned tokenizer trained with a novel Visual NTP objective. This closed-loop training ensures the visual tokens serve as a true semantic language compatible with LLM autoregression, overcoming the limitations of reconstruction-based tokenizers.

%     \item State-of-the-art Performance: We demonstrate that VoT achieves superior performance in both semantic alignment and complex reasoning-based generation tasks, validating the effectiveness of explicit visual thought modeling.
% \end{itemize}

\begin{itemize}
\item \textbf{Vision-of-Thought Framework.} We propose \emph{VoT}, a framework that inserts a discrete visual-thought branch between a VLM and a diffusion decoder, decoupling semantic reasoning from pixel-space rendering and improving cross-modal semantic alignment.

\item \textbf{VLM-Aligned VoT Tokenizer.} We introduce a tokenizer training recipe that quantizes \emph{VLM semantic features} rather than pixels and uses a closed-loop objective combining VLM alignment, feature reconstruction, and vector-quantization losses. We show that VLM alignment and reconstruction are complementary and that \emph{scaling the teacher VLM} consistently benefits token quality and downstream alignment.

\item \textbf{Efficient Unified MoT Integration and Strong Results.} We present a practical MoT-based instantiation of VoT by cloning VLM experts to initialize a VoT branch and freezing the VLM branch for efficient training and stable understanding. With a two-stage model-training pipeline after Stage 0 tokenizer training, using only text--image pairs throughout, we achieve substantial improvements in semantic alignment and reasoning-driven text-to-image generation.

\end{itemize}

% \begin{figure}[t!]
%     \includegraphics[width=1.0\linewidth]{figures/pipeline_compare.pdf}
%     \caption{Architecture.}
%     \label{fig:architecture}
% \end{figure}
\section{Related Work}
\subsection{Unified Multimodal Understanding and Generation}
The convergence of visual understanding and generation into unified architectures represents a pivotal frontier in AI research. Recent breakthroughs, particularly GPT-4o~\cite{openai2024gpt4o}, have demonstrated the potential of single models that can seamlessly interpret and generate multimodal content. Along this trajectory, various approaches have emerged to unify these capabilities. \textbf{Unified autoregressive models} such as Show-o~\cite{show-o}, Janus~\cite{chen2024janus}, and Unified-IO~\cite{lu2022unifiedio} attempt to handle both tasks within a single transformer architecture by treating images as discrete token sequences. However, these methods often compromise either generation fidelity or understanding depth because of the tension between reconstruction-oriented visual codes and semantics-oriented language modeling.

Alternatively, \textbf{diffusion-based unification efforts}~\cite{ge2023planting} typically graft understanding modules onto diffusion backbones, yet they suffer from a modality gap: the rich, structured knowledge of VLMs cannot be transferred effectively to diffusion decoders through static conditioning alone. VoT introduces an explicit, interpretable intermediate layer---discrete visual thoughts---that decouples semantic planning from pixel rendering, preserving the specialized strengths of both VLMs and diffusion transformers.

\subsection{Visual Tokenization Strategies}
Visual tokenization serves as the critical interface between continuous visual signals and discrete language-like representations. Early works such as VQ-VAE~\cite{oord2017neural} and VQGAN~\cite{esser2021taming} established the paradigm of compressing images into discrete tokens via reconstruction objectives. Subsequent methods such as LFQ~\cite{yu2024language} and MaskGIT~\cite{chang2022maskgit} improved codebook efficiency and generation diversity. However, these tokenizers remain fundamentally \textit{reconstruction-centric}: they optimize for pixel-level fidelity, producing ``compressed pixel fragments'' rather than semantic units.

Recent efforts to align visual tokens with language semantics have explored alternative pathways. SEED~\cite{ge2023planting} and LaVIT~\cite{jin2023unified} align visual tokens with language-model embeddings but focus primarily on post-hoc alignment rather than native integration into the VLM's reasoning process. Emu~\cite{sun2023emu} introduces a visual decoder for VLM-generated embeddings but relies on continuous rather than discrete representations. Across these approaches, the key limitation is the lack of closed-loop training that preserves visual information while making discrete tokens semantically readable by a VLM. Our VoT tokenizer addresses this gap with a unified objective that combines VLM alignment, feature reconstruction, and vector-quantization losses.

% \subsection{Autoregressive Image Generation}
% Autoregressive (AR) approaches to image generation have evolved from early pixel-level prediction \cite{oord2016pixel} to modern discrete token-based methods. DALL-E \cite{ramesh2021zero} and CogView \cite{ding2021cogview} demonstrated the feasibility of training large transformers on VQ-VAE tokens, but suffered from the \textit{representation mismatch} problem: tokens optimized for reconstruction exhibit long-tailed distributions and weak semantic correlations, making them inherently unstable for AR modeling.
% Recent works have explored hybrid approaches. LlamaGen \cite{sun2024llamagen} treats image tokens as a foreign language for LLMs, yet relies on standard VQ-tokenizers without semantic alignment. More relevant to our approach are methods that leverage pre-trained VLMs for generation. \textbf{Query-based approaches} such as MetaQuery \cite{dong2023metaquery} and BLIP-3 \cite{li2024blip3} generate parallel visual queries to condition diffusion models, offering computational efficiency but sacrificing the structural consistency required for complex reasoning. In contrast, VoT adopts an autoregressive token generation strategy, enabling \textit{visual chain-of-thought}—a step-by-step compositional planning capability that parallel query generation cannot achieve. This distinction is crucial: while queries provide parallel conditioning, VoT tokens provide sequential reasoning, effectively extending the LLM's cognitive architecture into the visual domain.

\subsection{Reasoning and Planning in Visual Generation}
The integration of reasoning capabilities into generative models has gained significant traction. Text-to-image systems such as DALL-E 3~\cite{betker2023improving} and Imagen~\cite{saharia2022photorealistic} improve prompt following through sophisticated text encoding yet lack explicit intermediate reasoning steps. Recent efforts to introduce chain-of-thought (CoT) mechanisms into vision tasks~\cite{hu2024visual,zhang2024multimodal} have primarily focused on understanding rather than generation.

Several concurrent works explore planning-based generation. LayoutGPT~\cite{wu2024layoutgpt} leverages LLMs for layout planning before image synthesis. However, these methods typically treat planning and generation as separate stages. VoT integrates planning \textit{within} the generative process through a unified token vocabulary. By generating discrete visual plans as VoT tokens, our framework enables the VLM to perform ``visual thinking''---reasoning about composition, spatial relationships, and attributes before committing to pixel values. This shifts the paradigm from direct \textit{text-to-image mapping} to \textit{text-to-thought-to-image generation}, providing interpretability and controllability that are absent from end-to-end diffusion or query-based approaches.
\section{Overview of VoT}

\begin{figure*}[t!]
    \centering
    \includegraphics[width=0.9\linewidth]{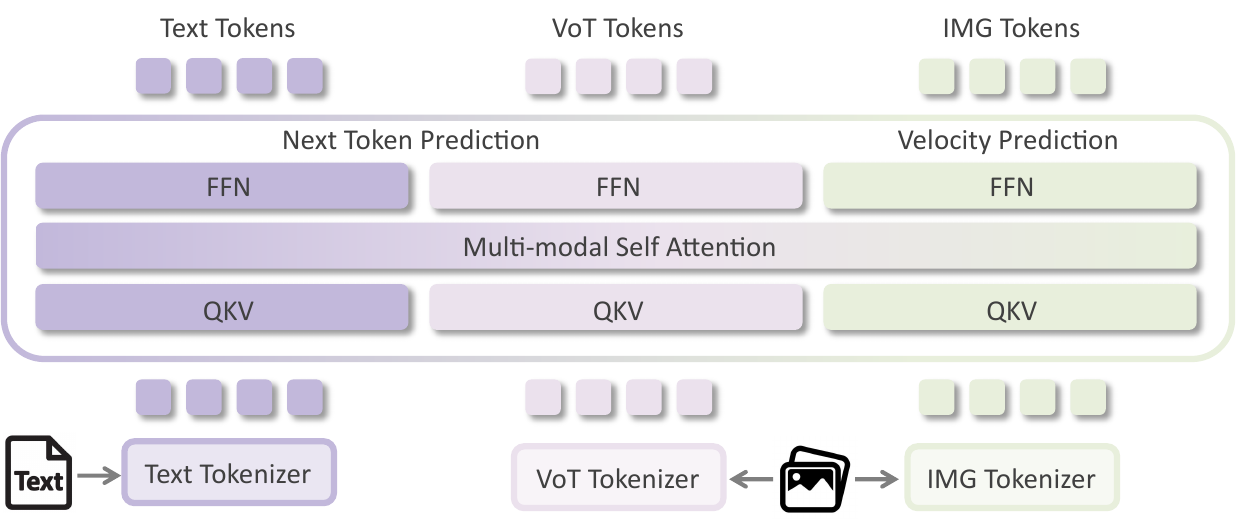}
    \caption{Architecture of the unified vision-language model with VoT. We extend MoT to three branches: \emph{Text}, \emph{VoT}, and \emph{Diffusion}. Key--value pairs from all branches are concatenated for unified attention. For text-to-image and image-to-image generation, the VoT branch first predicts discrete visual tokens as semantic plans; the diffusion branch then attends jointly to the text and VoT key--value pairs to synthesize the final image.}
    \label{fig:pipeline}
\end{figure*}

% Our proposed Vision-of-Thought (VoT) framework aims to bridge the modality gap between high-level semantic understanding and low-level visual synthesis. Instead of mapping text directly to pixel space, we introduce a discrete VoT representation as an intermediate semantic interface. This design allows us to leverage the reasoning capabilities of Large Vision-Language Models (VLMs) to generate structured visual plans, which are subsequently translated into high-fidelity images by a diffusion model. By decoupling the generation process into discrete semantic planning and continuous perceptual rendering, VoT ensures better cross-modal alignment and enables the VLM to act as a unified reasoning backbone.

% The training framework is structured into three progressive stages. First, in Stage 1: VoT Tokenizer Training, we construct a discrete visual codebook aligned with the VLM's semantic space. Unlike standard reconstruction-based tokenizers, we incorporate a Visual Next-Token Prediction (NTP) objective to ensure the generated tokens are statistically compatible with the VLM's autoregressive mechanism. Second, in Stage 2: Autoregressive Visual Generation, we freeze the VLM backbone and train lightweight experts to autoregressively generate VoT token sequences from text prompts, effectively transferring the VLM's planning capabilities to the visual domain. Finally, in Stage 3: VoT-Conditioned Diffusion Training, we train a diffusion transformer conditioned on these discrete VoT tokens to synthesize the final high-quality images.

Our proposed \emph{Vision-of-Thought (VoT)} framework bridges the gap between semantic understanding and visual synthesis by inserting a discrete visual-thought layer between a VLM and a diffusion decoder. Rather than mapping text directly to pixels, the VLM first produces structured VoT tokens that act as explicit visual plans, which are then rendered by the diffusion branch in the unified model (Fig.~\ref{fig:pipeline}).

\begin{figure}[t!]
    \centering
    \includegraphics[trim={1cm 0.5cm 1cm 0.5cm}, clip, width=0.9\linewidth]{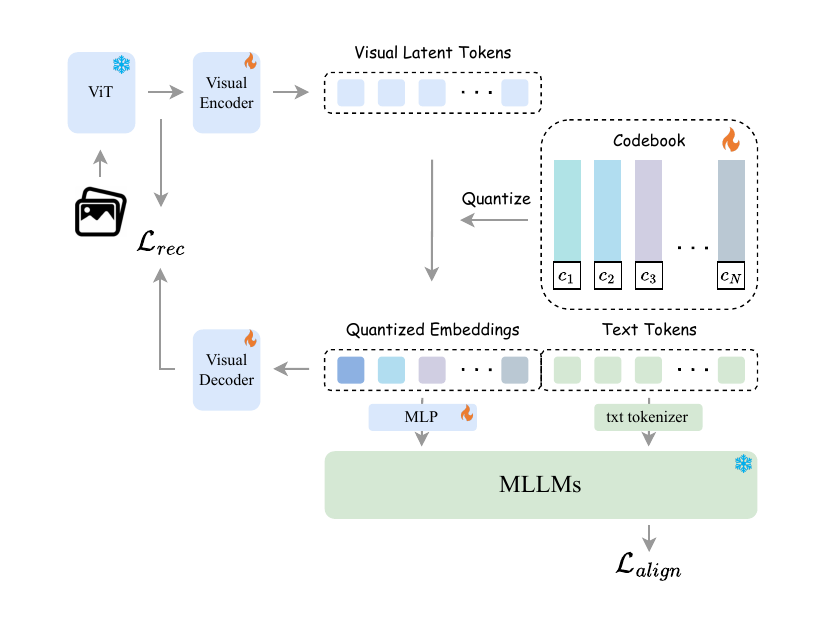}
    \caption{VLM-aligned VoT tokenizer training pipeline. Images are encoded by a frozen VLM ViT, adapted by a trainable encoder, and discretized with SimVQ. Training combines (1) feature reconstruction loss $\mathcal{L}_{rec}$, (2) VLM alignment loss $\mathcal{L}_{align}$, and (3) vector-quantization loss $\mathcal{L}_{VQ}$.}
    \label{fig:tokenizer}
\end{figure}

The framework follows three stages. In \emph{Stage 0: VLM-Aligned VoT Tokenizer Training} (Sec.~\ref{sec:tokenizer}, Fig.~\ref{fig:tokenizer}), we learn a discrete codebook in the VLM semantic space using feature reconstruction, VLM alignment, and vector-quantization losses. In \emph{Stage 1: VoT Branch Pretraining} (Sec.~\ref{sec:unified}), we freeze the pretrained VLM and diffusion branches and train lightweight VoT experts to predict visual tokens from text prompts using targets extracted from paired images. In \emph{Stage 2: VoT-DiT Joint Training} (Sec.~\ref{sec:unified}), we jointly fine-tune the VoT and diffusion branches while keeping the VLM branch frozen to preserve multimodal understanding and reduce training cost.

\section{VLM-Aligned VoT Tokenizer Training}
\label{sec:tokenizer}

The cornerstone of our framework is a discrete visual representation natively compatible with a vision-language model (VLM). Unlike traditional tokenizers prioritizing pixel-level fidelity, our \emph{VoT Tokenizer} minimizes the distributional shift between visual tokens and the VLM's semantic space via a specialized architecture and a unified alignment objective.

\subsection{Semantically Aligned Tokenizer Architecture}

Our architecture (Fig.~\ref{fig:tokenizer}) is designed to fully harness the pretrained capabilities of state-of-the-art VLMs. The pipeline comprises three primary modules:

\paragraph{Foundation \& Adapter Encoder.} The encoder integrates a frozen foundation model with a trainable adapter. We denote the Vision Transformer (ViT) of the target VLM (Qwen2.5-VL) as $\mathcal{E}_{\text{frozen}}$, keeping it \textbf{frozen} to preserve its robust, pre-aligned semantic features. To bridge the modality gap between continuous ViT features and the discrete latent space, we append a lightweight trainable adapter $\mathcal{E}_{\text{adapt}}$ (a 1-layer ViT block, denoted as \emph{Visual Encoder} in Fig.~\ref{fig:tokenizer}) to the backbone. Given an input image $\mathbf{x}$, adapted semantic features $\mathbf{h}$ are derived via:
\begin{equation}
    \mathbf{h} = \mathcal{E}_{\text{adapt}}(\mathcal{E}_{\text{frozen}}(\mathbf{x})).
\end{equation}
This adapter projects foundation features onto a manifold suitable for quantization while retaining the VLM's semantic structure.

\paragraph{SimVQ Head.} We employ a Similarity-based Vector Quantization (SimVQ) mechanism with a learnable codebook $\mathcal{C} = \{c_k\}_{k=1}^K \subset \mathbb{R}^d$. The discrete code index $k$ for a feature vector $\mathbf{h}_i$ is selected by minimizing the Euclidean distance:
\begin{equation}
    k = \mathop{\arg\min}_{j \in \{1, \dots, K\}} \| \mathbf{h}_i - c_j \|_2^2.
\end{equation}
To prevent codebook collapse, we regularize the similarity logits $s_{i,j} = -\| \mathbf{h}_i - c_j \|_2^2$ with an \textbf{entropy loss}. The quantized feature vector $\mathbf{z}_q$ is passed to subsequent layers, with gradients propagated via the straight-through estimator: $\mathbf{z}_{st} = \mathbf{h} + \text{sg}[\mathbf{z}_q - \mathbf{h}]$, where $\text{sg}[\cdot]$ denotes the stop-gradient operator.

\paragraph{Deep Semantic Decoder.}
The decoder $\mathcal{D}$ reconstructs continuous semantic features from quantized tokens $\mathbf{z}_{st}$ via $\hat{\mathbf{f}} = \mathcal{D}(\mathbf{z}_{st})$. Departing from typical lightweight designs, we employ a 6-layer transformer decoder. As shown in Table~\ref{tab:ablation_settings}, increasing the decoder depth from 2 layers in Exp.D (69.32) to 6 layers in Exp.E (70.37) consistently improves AutoEval. We attribute this gain to the decoder's increased capacity to \emph{invert quantization} in the teacher feature space: a stronger decoder reduces the pressure on the codebook to preserve fine-grained details, allowing tokens to focus on language-grounded semantics while the decoder smooths quantization noise when reconstructing dense ViT features.

% \subsection{Training Objectives}

% Our training paradigm integrates three complementary forces: visual grounding, semantic readability (understanding), and generative compatibility (planning).

% \paragraph{Feature Reconstruction Loss ($\mathcal{L}_{rec}$): Visual Grounding.} To prevent the codebook from collapsing into abstract concepts without visual discriminability, we enforce a reconstruction loss on the ViT features. This acts as the visual anchor:
% \begin{equation}
%     \mathcal{L}_{rec} = \| \mathcal{E}_{\text{frozen}}(\mathbf{x}) - \mathcal{D}(\mathbf{z}_{st}) \|_2^2
% \end{equation}

% \paragraph{VLM alignment Loss ($\mathcal{L}_{align}$): Semantic Readability.} This objective ensures that the discretized tokens retain sufficient semantics to be ``understood'' by the VLM. We feed the quantized features (after VQ) into the frozen VLM and train the tokenizer to minimize the standard Cross-Entropy (CE) loss for image captioning.
% \begin{equation}
%     \mathcal{L}_{align} = - \sum_{i} \log P_{VLM}(\text{text}_i \mid \text{text}_{<i}, \mathbf{z}_{img})
% \end{equation}
% By optimizing this, we force the codebook to preserve the semantic information necessary for the VLM to describe the image content accurately.

% \paragraph{Total Objective.}
% \begin{equation}
%     \mathcal{L}_{Stage1} = \mathcal{L}_{rec} + \lambda_{align} \mathcal{L}_{align} + \mathcal{L}_{commit}
% \end{equation}

\subsection{Training Objectives}

We train the VoT tokenizer with three complementary objectives as follows. Throughout training, the teacher ViT encoder and the VLM are kept frozen.

\paragraph{Feature Reconstruction Loss ($\mathcal{L}_{rec}$): Visual Grounding.}
Given an image $\mathbf{x}$, the frozen teacher ViT produces token features
$\mathbf{f} = \mathcal{E}_{\text{frozen}}(\mathbf{x}) \in \mathbb{R}^{N \times d}$.
After quantization and straight-through estimation, we obtain $\mathbf{z}_{st}$, and
a semantic decoder $\mathcal{D}$ reconstructs teacher features
$\hat{\mathbf{f}} = \mathcal{D}(\mathbf{z}_{st})$.
We use a token-wise mean-squared error:
\begin{equation}
\mathcal{L}_{rec} = \frac{1}{N}\left\| \mathbf{f} - \hat{\mathbf{f}} \right\|_2^2.
\end{equation}

\paragraph{VLM Alignment Loss ($\mathcal{L}_{align}$): Semantic Readability.}
This objective enforces that the discretized representation remains readable by the frozen VLM.
We feed the tokenizer-produced visual embeddings (denoted as $\mathbf{v}$; e.g., $\mathbf{v}=\mathbf{z}_{st}$ or a projected version thereof) into the frozen VLM and minimize the standard language-model cross-entropy on the paired caption $\mathbf{y}=(y_1,\dots,y_T)$:
\begin{equation}
\mathcal{L}_{align} = - \sum_{t=1}^{T} \log P_{\text{VLM}}\!\left(y_t \mid y_{<t}, \mathbf{v}\right).
\end{equation}

\paragraph{Vector-Quantization Loss ($\mathcal{L}_{VQ}$): Stable Quantization.}
Let $\mathbf{h}$ denote the pre-quantization features (the tokenizer output before nearest-neighbor lookup),
and $\mathbf{z}_q$ denote the selected codebook embeddings.
We use the standard VQ objective (codebook loss + commitment loss):
\begin{equation}
\mathcal{L}_{VQ} =
\left\| \text{sg}[\mathbf{h}] - \mathbf{z}_q \right\|_2^2
+ \beta \left\| \mathbf{h} - \text{sg}[\mathbf{z}_q] \right\|_2^2,
\end{equation}
where $\text{sg}[\cdot]$ is the stop-gradient operator and $\beta$ is the commitment weight.

\paragraph{Total Objective.}
The full tokenizer training objective is:
\begin{equation}
\mathcal{L}_{tok} = \mathcal{L}_{rec} + \lambda_{align}\mathcal{L}_{align} + \lambda_{VQ}\mathcal{L}_{VQ}.
\end{equation}

\section{Unified Vision-Language Model with VoT}
\label{sec:unified}
Building upon our pretrained visual tokenizer, we now introduce a unified multimodal architecture that integrates discrete visual tokens into a joint understanding and generation framework. Our approach extends the Mixture-of-Transformers (MoT) paradigm to support three specialized pathways: multimodal understanding, visual token prediction, and diffusion-based image synthesis.

\subsection{Unified Three-Branch MoT Backbone}

Building upon the Mixture-of-Transformers (MoT) paradigm~\cite{deng2025emerging,liao2025mogao}, which harmonizes language and diffusion generation within a shared backbone, we extend the architecture to a three-branch design by incorporating a dedicated \emph{VoT branch} for discrete visual token modeling. In this unified framework, distinct modalities share the transformer backbone while using modality-specific components to preserve their unique properties. Specifically, for each transformer layer $\ell$, we maintain separate hidden states for text ($\mathbf{h}^{(\ell)}_{\text{txt}}$), visual tokens ($\mathbf{h}^{(\ell)}_{\text{vot}}$), and diffusion latents ($\mathbf{h}^{(\ell)}_{\text{diff}}$). Each branch $m \in \{\text{txt}, \text{vot}, \text{diff}\}$ applies independent layer normalization (LN) and projections to compute queries, keys, and values:
\begin{equation}
\begin{split}
\mathbf{Q}_m &= \mathbf{W}^Q_m \cdot \text{LN}_m(\mathbf{h}_m^{(\ell)}), \\
\mathbf{K}_m &= \mathbf{W}^K_m \cdot \text{LN}_m(\mathbf{h}_m^{(\ell)}), \\
\mathbf{V}_m &= \mathbf{W}^V_m \cdot \text{LN}_m(\mathbf{h}_m^{(\ell)}).
\end{split}
\end{equation}
To facilitate cross-modal information exchange, we construct a global context by concatenating the key--value pairs from all branches:
\begin{equation}
\mathbf{K} = [\mathbf{K}_{\text{txt}}; \mathbf{K}_{\text{vot}}; \mathbf{K}_{\text{diff}}], \quad
\mathbf{V} = [\mathbf{V}_{\text{txt}}; \mathbf{V}_{\text{vot}}; \mathbf{V}_{\text{diff}}].
\end{equation}
The unified attention mechanism then operates over this joint space, modulated by a branch-specific mask $\mathbf{M}_m$:
\begin{equation}
\text{Attn}_m = \text{softmax}\left(\frac{\mathbf{Q}_m \mathbf{K}^\top}{\sqrt{d_k}} + \mathbf{M}_m\right) \mathbf{V}.
\end{equation}
Crucially, we employ causal masking for the text and VoT branches to enable autoregressive generation, whereas the diffusion branch uses bidirectional attention. Finally, the layer output is obtained by processing the attention result through a branch-specific feed-forward network (FFN):
\begin{equation}
\mathbf{h}_m^{(\ell+1)} = \mathbf{h}_m^{(\ell)} + \mathbf{W}^O_m \cdot \text{Attn}_m + \text{FFN}_m(\text{LN}'_m(\cdot)).
\end{equation}
This architecture balances shared semantic modeling with modality-specific processing.

% \paragraph{Input Representations.}
% Text tokens are embedded via a shared vocabulary embedding $\mathbf{E}_{\text{txt}}$. For VOT, we introduce a \emph{Visual Token Embedding} (VTE) table $\mathbf{E}_{\text{vot}} \in \mathbb{R}^{K \times d_v}$, where $K = 65536$ corresponds to the codebook size of our pretrained visual tokenizer. An optional projection $\mathbf{W}_{\text{vte}} \in \mathbb{R}^{d_v \times d}$ aligns the VTE dimension with the transformer hidden dimension. For diffusion, continuous VAE latents are patchified and projected via a linear layer.

% \paragraph{Output Heads.}
% The text branch shares the language model head with the original LLM. For VOT prediction, we introduce a dedicated head $\mathbf{W}_{\text{vot\_head}} \in \mathbb{R}^{d \times K}$ that maps hidden states to logits over the visual codebook. The diffusion branch outputs velocity predictions for the denoising process.

% \paragraph{Advantages.}
% This three-branch design offers several benefits:
% \begin{itemize}
%     \item \textbf{Semantic grounding}: VOT provides a discrete, semantically meaningful intermediate representation that bridges text descriptions and pixel-level generation.
%     \item \textbf{Unified attention}: All modalities can attend to each other, enabling the diffusion branch to leverage both textual and visual token conditions simultaneously.
%     \item \textbf{Modular training}: The separation of expert parameters allows stage-wise training and efficient fine-tuning of individual branches.
% \end{itemize}

We implement VoT by adding per-layer VoT experts and a codebook-sized embedding and prediction interface for visual tokens, while the diffusion branch predicts denoising velocities. To integrate VoT efficiently into a unified MoT backbone, we initialize the VoT experts by copying a subset of pretrained VLM experts and keep the VLM branch frozen to preserve multimodal understanding and reduce training cost. At inference time, we first generate VoT tokens autoregressively conditioned on text, then render the final image with the diffusion branch conditioned on both text and VoT KV caches.

\subsection{Training Recipe}
Following Stage 0 tokenizer training, we adopt a two-stage strategy to integrate the VoT branch into the pretrained Mogao model.
% VOT AR
\paragraph{Stage 1: VoT Branch Pretraining.}
In this stage, we warm up the newly introduced VoT branch while freezing the pretrained backbone, so that VoT learns to represent and predict discrete visual tokens without perturbing the model's existing understanding and rendering capabilities.

We initialize the backbone from \emph{Mogao}~\cite{liao2025mogao}, a 14B MoT-based unified model. We load the pretrained weights for the VLM and diffusion branches and initialize the per-layer VoT experts by copying a subset of pretrained VLM experts. Only the remaining VoT-specific components are randomly initialized (std.\ 0.0125): the visual token embedding table $\mathbf{E}_{\mathrm{vot}} \in \mathbb{R}^{65{,}536 \times 2{,}048}$ and the VoT prediction head. We use the frozen visual tokenizer from Stage 0 (a Qwen2.5-VL ViT encoder followed by an adapter encoder and a SimVQ quantizer; Sec.~\ref{sec:tokenizer}) to extract discrete supervision targets and keep it fixed throughout this stage. The training objective is an autoregressive cross-entropy loss over a 65,536-way codebook, predicting 1,024 VoT tokens per image.

\paragraph{Stage 2: VoT-DiT Joint Training.}
Starting from the Stage 1 checkpoint, we jointly train the \emph{VoT branch} and the \emph{diffusion branch} so that they can co-adapt for high-quality synthesis. Concretely, we initialize from the warmed-up VoT model and update parameters in both pathways, including VoT experts, diffusion experts, and shared components. This joint optimization allows the diffusion branch to better exploit VoT conditioning, while the VoT branch refines its token representations to be more useful for generation.

We optimize a combined objective consisting of the \emph{VoT autoregressive prediction loss} and the \emph{diffusion velocity prediction loss}. Unlike Stage 1, where gradients are restricted to VoT modules, the diffusion loss now backpropagates through the shared attention layers and updates both branches, tightening the coupling between token prediction and rendering quality.

To support classifier-free guidance at inference time, we apply structured conditional dropout during training: VoT tokens are dropped with probability $0.5$, and text tokens are dropped with probability $0.1$ when VoT is present, resulting in approximately $50\%$ text-only, $45\%$ text-plus-VoT, and $5\%$ VoT-only samples. In addition, we randomly mask $0$--$20\%$ of VoT key--value pairs during attention, which prevents the diffusion branch from over-relying on VoT and improves robustness to incomplete conditions.

% \paragraph{Stage 3: Post-Training and RL}
% 高分辨率图像的训练, post train, RL
\section{Experiments}
\label{sec:experiment}
\subsection{Implementation Details}

% Stage 1: Training is conducted on 64 GPUs with a maximum sequence length of 24,576 tokens, using a linear warmup over 1,000 iterations. We train for approximately 12,000 iterations until the VOT prediction accuracy stabilizes.

% Stage 2: Training uses the AdamW optimizer with learning rate $2 \times 10^{-5}$, $\beta = (0.9, 0.95)$, and gradient clipping at max norm 1.0. We adopt a longer warmup of 2,500 iterations to ensure stable joint optimization given the increased number of trainable parameters. The maximum sequence length remains 24,576 tokens, and training is distributed across 64 GPUs with bf16 mixed precision.

\paragraph{Data Configuration.}
In Stage 0, we use image--text pairs with images processed at native aspect ratios and resolutions from approximately $434 \times 434$ to $1024 \times 1024$ pixels (188K--1M total pixels). We set the maximum sequence length to 16,384 tokens, with a per-sample limit of 12,288 tokens. Stages 1 and 2 use a large-scale text-to-image dataset. Additional dataset statistics and training configurations are provided in the Appendix.

\paragraph{Model Configuration.} We use the Qwen2.5-VL-7B ViT as the frozen vision encoder for the visual tokenizer. The SimVQ quantizer comprises a 1-layer encoder, a codebook with 65,536 entries ($d=128$), and a 6-layer decoder. We train it with a learning rate of $3 \times 10^{-4}$ using vector-quantization, feature reconstruction, and VLM alignment losses. For the unified vision-language model, we initialize from the pretrained Mogao-14B checkpoint and add a dedicated 7B-parameter VoT branch, yielding 21B parameters in total.

% \paragraph{Training Configuration.}
% All stages are trained on 64 GPUs using AdamW optimizer with $\beta = (0.9, 0.95)$ and gradient clipping at max norm 1.0. We use bf16 mixed precision and FlashAttention-3 for computational efficiency. Table~\ref{tab:training_config} summarizes the key configurations.

% \begin{table}[h]
% \centering
% \small
% \begin{tabular}{lccc}
% \toprule
% & \textbf{Stage-0} & \textbf{Stage-I} & \textbf{Stage-II} \\
% \midrule
% Trainable modules & Tokenizer & VOT branch & All \\
% Base model & Qwen2.5-VL-7B & Mogao-14B & Stage-I ckpt \\
% Learning rate & $3 \times 10^{-4}$ & -- & $2 \times 10^{-5}$ \\
% Warmup iterations & 2,500 & 1,000 & 2,500 \\
% Max sequence length & 16,384 & 24,576 & 24,576 \\
% Image resolution & $\sim$450$^2$ & $256^2$ & $256^2$ \\
% \bottomrule
% \end{tabular}
% \caption{Training configurations for each stage.}
% \label{tab:training_config}
% \end{table}

\subsection{Text-to-Image Generation}

\definecolor{myblue}{rgb}{0.93, 0.95, 0.99}
\definecolor{myred}{rgb}{0.98, 0.92, 0.92}
\begin{table*}[t]
    \centering
    \setlength{\tabcolsep}{4pt} % Adjust padding to fit
    \renewcommand{\arraystretch}{1.2}
    \scriptsize
    \caption{\textbf{Quantitative evaluation on GenEval benchmark.} We compare VoT with state-of-the-art ``Gen. Only'' (generation-specialized) and ``Unified'' (understanding + generation) models. $\dagger$ denotes methods utilizing an LLM rewriter. \textbf{VoT} achieves the highest overall score on GenEval, showing significant advantages in spatial and counting tasks.}
    \label{tab:geneval}
    \begin{tabular}{clccccccc}
        \toprule
        \textbf{Type} & \textbf{Model} & \textbf{Single Obj.} & \textbf{Two Obj.} & \textbf{Counting} & \textbf{Colors} & \textbf{Position} & \textbf{Color Attri.} & \textbf{Overall$\uparrow$} \\
        \midrule
        \multirow{8}{*}{\rotatebox{90}{\textit{Gen. Only}}}
        & PixArt-$\alpha$~\cite{chen2023pixart} & 0.98 & 0.50 & 0.44 & 0.80 & 0.08 & 0.07 & 0.48 \\
        & SDv$2.1$~\cite{rombach2022high} & 0.98 & 0.51 & 0.44 & 0.85 & 0.07 & 0.17 & 0.50 \\
        & DALL-E $2$~\cite{dalle2}  & 0.94 & 0.66 & 0.49 & 0.77 & 0.10 & 0.19 & 0.52 \\
        & Emu$3$-Gen~\cite{emu3}  & 0.98 & 0.71 & 0.34 & 0.81 & 0.17 & 0.21 & 0.54 \\
        & SDXL~\cite{sdxl} & 0.98 & 0.74 & 0.39 & 0.85 & 0.15 & 0.23 & 0.55 \\
        & DALL-E $3$~\cite{dalle3} & 0.96 & 0.87 & 0.47 & 0.83 & 0.43 & 0.45 & 0.67 \\
        & SD3-Medium~\cite{SD3} & 0.99 & 0.94 & 0.72 & 0.89 & 0.33 & 0.60 & 0.74 \\
        & FLUX.1-dev$^{\dagger}$~\cite{flux} & 0.98 & 0.93 & 0.75 & 0.93 & 0.68 & 0.65 & 0.82 \\
        \midrule
        \multirow{14}{*}{\rotatebox{90}{\textit{Unified}}}
        & Chameleon~\cite{chameleon} & -- & -- & -- & -- & -- & -- & 0.39 \\
        & LWM~\cite{lwm} & 0.93 & 0.41 & 0.46 & 0.79 & 0.09 & 0.15 & 0.47 \\
        & SEED-X~\cite{seed-x}  & 0.97 & 0.58 & 0.26 & 0.80 & 0.19 & 0.14 & 0.49 \\
        & TokenFlow-XL~\cite{qu2024tokenflow} & 0.95 & 0.60 & 0.41 & 0.81 & 0.16 & 0.24 & 0.55 \\
        & ILLUME~\cite{wang2024illume} & 0.99 & 0.86 & 0.45 & 0.71 & 0.39 & 0.28 & 0.61 \\
        & Janus~\cite{chen2024janus} & 0.97 & 0.68 & 0.30 & 0.84 & 0.46 & 0.42 & 0.61 \\
        & Transfusion~\cite{zhou2024transfusion} & -- & -- & -- & -- & -- & -- & 0.63 \\
        & Emu$3$-Gen$^{\dagger}$~\cite{emu3} & 0.99 & 0.81 & 0.42 & 0.80 & 0.49 & 0.45 & 0.66 \\
        & Show-o~\cite{show-o} & 0.98 & 0.80 & 0.66 & 0.84 & 0.31 & 0.50 & 0.68 \\
        & Janus-Pro-7B~\cite{januspro2025} & 0.99 & 0.89 & 0.59 & 0.90 & 0.79 & 0.66 & 0.80 \\
        & MetaQuery-XL$^{\dagger}$~\cite{pan2025transfer} & -- & -- & -- & -- & -- & -- & 0.80 \\
        & BAGEL~\cite{deng2025emerging} & 0.99 & 0.94 & 0.81 & 0.88 & 0.64 & 0.63 & 0.82 \\
        & Mogao~\cite{liao2025mogao} & 1.00 & 0.97 & 0.83 & 0.93 & 0.84 & 0.80 & 0.89 \\
        \rowcolor{myblue} 
        & \textbf{VoT (Ours)} & 1.00 & 0.98 & 0.86 & 0.95 & 0.85 & 0.82 & \textbf{0.91} \\
        \bottomrule
    \end{tabular}
\end{table*}

\begin{figure}[t!]
    \centering
    \includegraphics[width=0.9\linewidth]{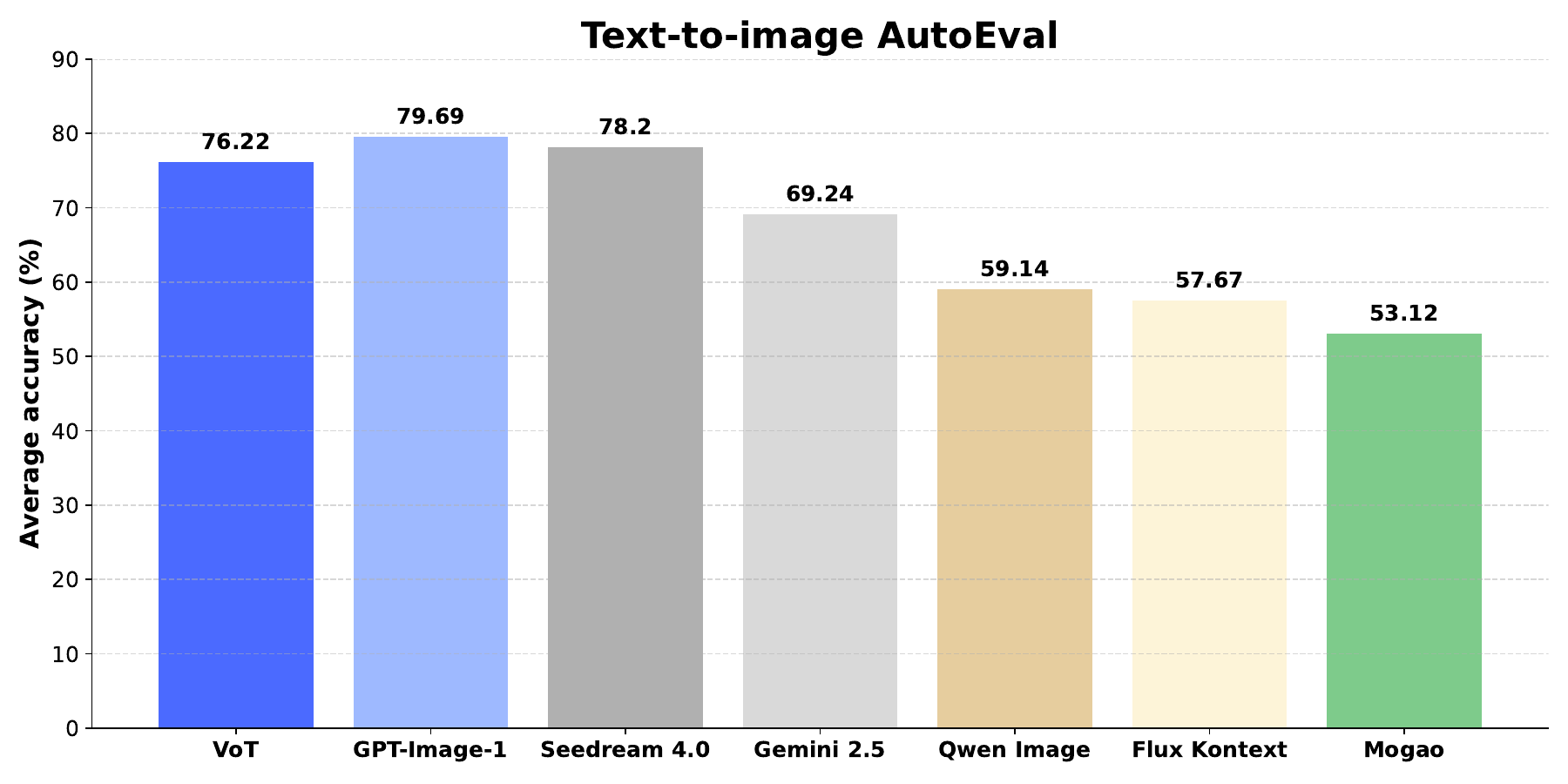}
    \caption{Evaluation of text-to-image instruction alignment.}
    \label{fig:auto_eval}
\end{figure}

\paragraph{Evaluation Setup.} We evaluate text-to-image generation on the public GenEval benchmark~\cite{ghosh2023geneval} and an internal benchmark. GenEval measures compositional text-to-image alignment with object detectors, covering object categories, counts, colors, and spatial relations. It contains 553 prompts, and we generate four images per prompt. We also evaluate on our internal DreamBench suite with an automatic pipeline and report its aggregate metric as AutoEval.

\paragraph{Quantitative Comparison.}
We report GenEval results in Table~\ref{tab:geneval}. VoT achieves an overall score of 0.91, outperforming leading generation-specialized models (e.g., FLUX.1-dev, 0.82) and recent unified multimodal models (e.g., BAGEL, 0.82; Mogao, 0.89). VoT also shows clear advantages on compositionally demanding tasks such as counting (0.86 vs.\ 0.83 for Mogao). These results support our hypothesis that the explicit discrete VoT layer transfers the VLM's planning and reasoning capabilities to the generation process, helping satisfy structural constraints that implicit conditioning methods often fail to handle.

On our internal instruction-following benchmark (Fig.~\ref{fig:auto_eval}), VoT achieves strong prompt--image coherence and outperforms baselines such as Seedream 4.0 and GPT-Image-1, demonstrating the benefit of VoT tokens for instruction adherence.

\begin{figure}[t!]
    \centering
    \includegraphics[width=1.0\linewidth]{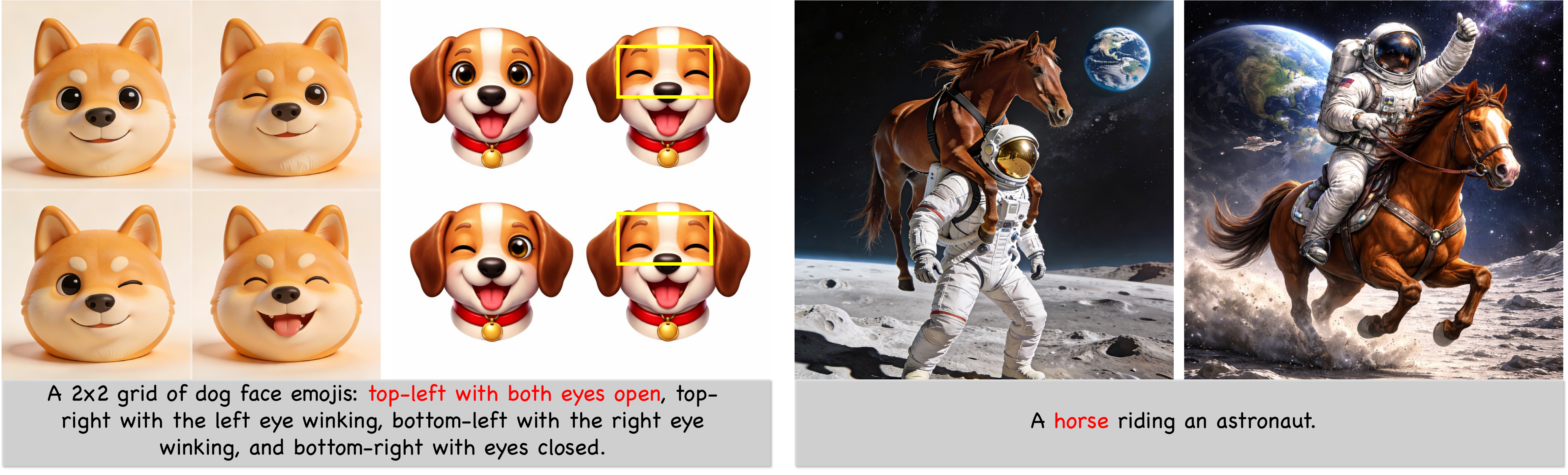}
    \caption{Text-to-image comparison against GPT-Image-1. For each pair of images, left: VoT; right: GPT-Image-1.}
    \label{fig:visual_comp}
\end{figure}

\paragraph{Visual Comparison.}
To qualitatively assess model capabilities in challenging scenarios, we select representative cases involving complex layout control and counterfactual generation, as shown in Fig.~\ref{fig:visual_comp}. We compare our results with GPT-Image-1. Our model (left) accurately renders precise attributes (e.g., facial expressions) and surreal scenes (e.g., ``a horse riding an astronaut''), whereas GPT-Image-1 often fails because of attribute leakage or reliance on training-distribution biases.

\subsection{Ablation Study}

% \begin{figure}[t!]
%     \includegraphics[width=1.0\linewidth]{figures/vlm_loss_comparison_v2.png}
%     \caption{VLM Alignment Loss during tokenizer training.}
%     \label{fig:loss_vlm}
% \end{figure}

% \begin{figure}[t!]
%     \includegraphics[width=1.0\linewidth]{figures/vot_arloss_comparison_v2.png}
%     \caption{AR Loss of VoT tokens in Stage 1.}
%     \label{fig:loss_ar}
% \end{figure}

\begin{figure}[t!]
    \centering
    %--- 第一个子图 (左侧) ---
    \begin{subfigure}[b]{0.495\linewidth}
        \centering
        \includegraphics[width=\linewidth]{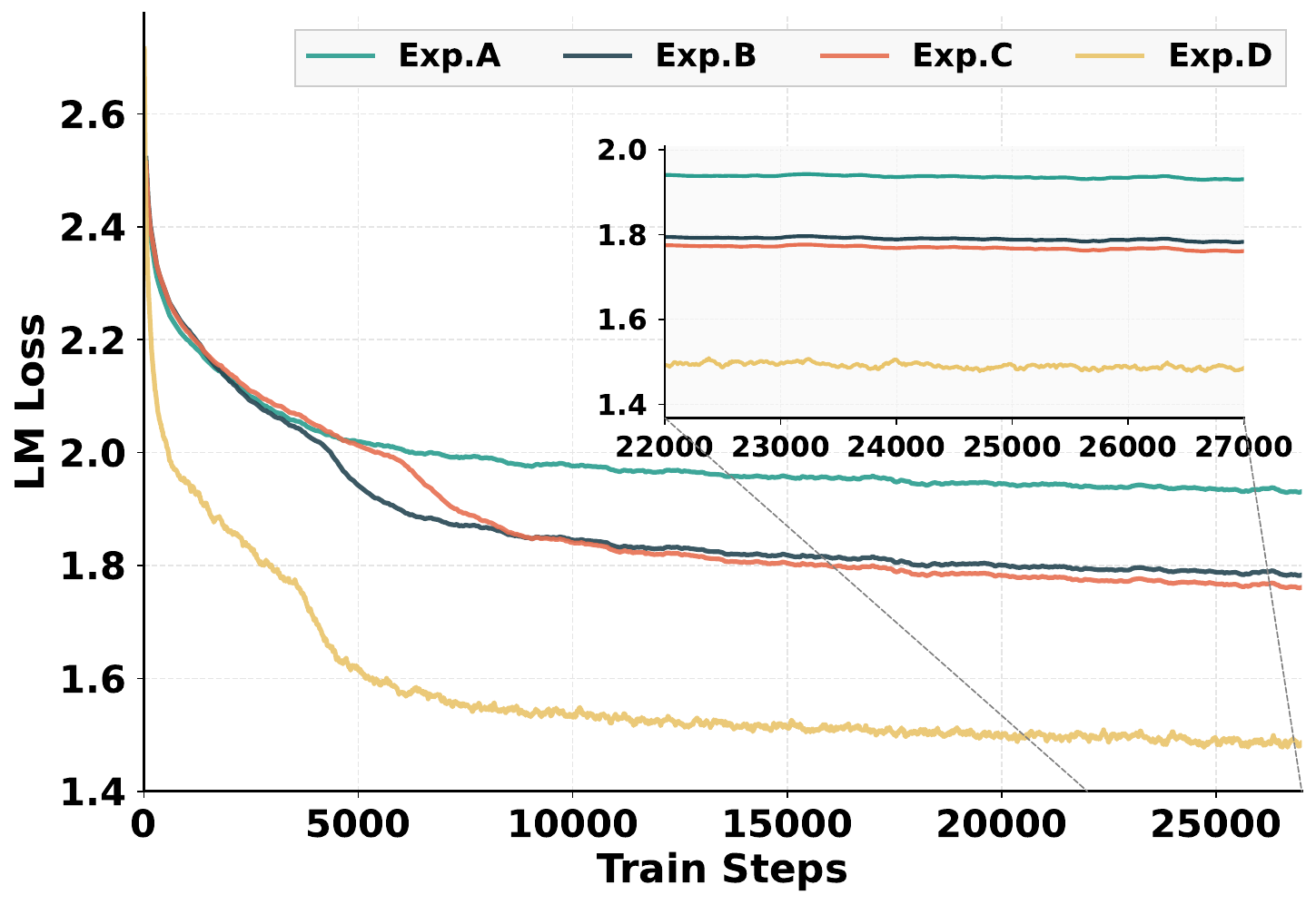}
        \caption{VLM alignment loss}
        \label{fig:vlm_loss_sub}     % 第一个小图的唯一标签
    \end{subfigure}
    \hfill % 在两个图之间添加弹性空白，使它们分居两侧
    %--- 第二个子图 (右侧) ---
    \begin{subfigure}[b]{0.495\linewidth}
        \centering
        \includegraphics[width=\linewidth]{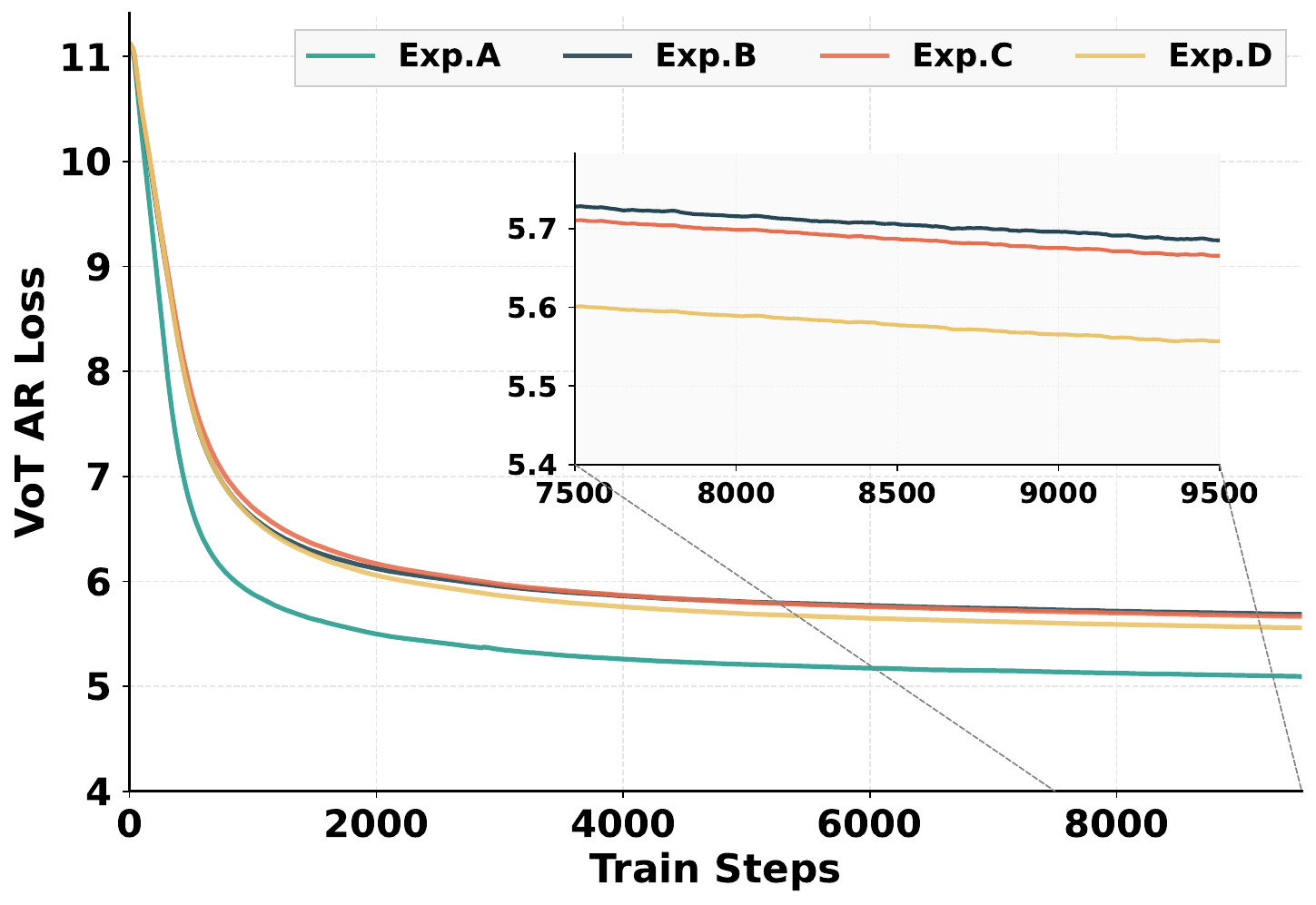}
        \caption{VoT autoregressive loss}
        \label{fig:loss_ar_sub}      % 第二一个小图的唯一标签
    \end{subfigure}
    %--- 整个大图的总标题和总标签 ---
    \caption{Training curves for the VLM alignment loss (left) and VoT autoregressive loss (right).}
    \label{fig:loss_comparisons_combined}
\end{figure}

To validate each component of VoT tokenizer training, we evaluate six configurations (A--F), as summarized in Table~\ref{tab:ablation_settings}. Configurations A--D isolate the effects of training objectives, architectural adaptation, and teacher scaling; E and F evaluate the full VoT model before and after post-training.

\begin{table}[htbp]
\centering
\small
\caption{Ablation study of tokenizer configurations. We progressively integrate reconstruction loss, trainable adapters, and teacher scaling. \textbf{Enc/Dec}: trainable encoder/decoder layers. \textbf{Align/Rec}: VLM alignment/reconstruction loss. All tokenizer variants use $\mathcal{L}_{VQ}$. ${\dagger}$ denotes the final model after supervised fine-tuning (SFT) and reinforcement learning from human feedback (RLHF).}
\label{tab:ablation_settings}
\resizebox{0.9\linewidth}{!}{
\begin{tabular}{@{}lccccc@{}}
\toprule
\textbf{Exp.} & \textbf{Teacher VLM} & \textbf{Enc. / Dec. Depth} & \textbf{$\mathcal{L}_{align}$} & \textbf{$\mathcal{L}_{rec}$} & \textbf{AutoEval} \\ \midrule
\rowcolor{myred}
Mogao (baseline) & -- & -- / -- & -- & -- & 53.12 \\
A & Qwen2-VL 2B & -- / -- & \checkmark & -- & 58.67 \\
B & Qwen2-VL 2B & -- / 2 & \checkmark & \checkmark & 60.98 \\
C & Qwen2-VL 2B & 1 / 2 & \checkmark & \checkmark & 64.64 \\
D & Qwen2.5-VL 7B & 1 / 2 & \checkmark & \checkmark & 69.32 \\
\rowcolor{myblue}
E (VoT) & Qwen2.5-VL 7B & 1 / 6 & \checkmark & \checkmark & \textbf{70.37} \\
\rowcolor{myblue}
F (VoT${\dagger}$) & Qwen2.5-VL 7B & 1 / 6 & \checkmark & \checkmark & \textbf{76.22} \\ \bottomrule
\end{tabular}
}
\end{table}
\FloatBarrier

\paragraph{Benefits of Reconstruction and Adapters.}
In Fig.~\ref{fig:vlm_loss_sub}, LM loss drops from $\sim$1.93 (Exp.A) to $\sim$1.79 with reconstruction (Exp.B) and $\sim$1.76 with a 1-layer adapter (Exp.C). These components stabilize token learning and reduce the mismatch with frozen ViT features; AutoEval correspondingly rises from $58.67 \to 60.98 \to 64.64$.

\paragraph{VoT AR Loss vs. Tokenizer Quality.}
Fig.~\ref{fig:loss_ar_sub} shows the VoT autoregressive loss in Stage 1 (Sec.~\ref{sec:unified}). Exp.A achieves the lowest final autoregressive loss ($\sim$5.08) yet has the highest tokenizer LM loss in Fig.~\ref{fig:vlm_loss_sub} ($\sim$1.93). AutoEval further confirms that predictability is not synonymous with utility: despite its lower autoregressive loss, Exp.A obtains only 58.67, substantially below Exp.B and Exp.C. Thus, reconstruction and the adapter can slightly increase autoregressive difficulty by preserving more information while improving semantic alignment and downstream utility.

\paragraph{Effect of Teacher VLM Scale.}
Comparing Exps. A--C (Qwen2-VL 2B teacher) with Exp.D (Qwen2.5-VL 7B teacher), Fig.~\ref{fig:vlm_loss_sub} shows the LM loss dropping from $\sim$1.76--1.93 to $\sim$1.48, while AutoEval increases from 64.64 to 69.32. Although reconstruction can make autoregressive prediction harder by preserving more information (Fig.~\ref{fig:loss_ar_sub}), Exp.D still achieves the lowest VoT autoregressive loss among the reconstruction-based settings. This result suggests that a stronger teacher provides cleaner semantic supervision and induces a more structured, autoregressive-friendly visual lexicon.

\paragraph{Benefits of VoT and Post-Training.}
We next compare the original \emph{Mogao} baseline with our VoT model. Exp.E is the model after training with VoT tokens, making it the most direct comparison to \emph{Mogao} under the same T2I training data. As shown in Table~\ref{tab:ablation_settings}, VoT tokens yield a clear improvement in AutoEval (53.12 $\rightarrow$ 70.37). Exp.F further improves the score after SFT and RLHF, indicating that post-training is complementary to VoT and translates better planning into higher final generation quality.

\section{Conclusion}
In this work, we introduced \emph{Vision-of-Thought (VoT)}, a unified framework that bridges semantic understanding and visual synthesis by inserting a discrete visual planning layer between VLMs and diffusion decoders. By leveraging the VLM to generate structured VoT tokens as explicit plans before rendering, our approach successfully decouples high-level reasoning from pixel-level generation, significantly enhancing cross-modal alignment and interpretability. We hope VoT serves as a foundational step towards next-generation multimodal agents capable of coherent visual thinking and creation.

% Our key technical contribution is a novel training strategy for a \emph{semantically aligned tokenizer}, which demonstrates that reconstruction and semantic alignment are complementary objectives, and that scaling the VLM teacher substantially improves token quality. Implemented via an elegant MoT architecture, our method achieves state-of-the-art generation performance competitive with leading models like GPT-4o, while maintaining training efficiency by freezing the VLM backbone. We hope VoT serves as a foundational step towards next-generation multimodal agents capable of coherent visual thinking and creation.

\bibliographystyle{plainnat}
\bibliography{main}

\begin{thebibliography}{36}
\providecommand{\natexlab}[1]{#1}
\providecommand{\url}[1]{\texttt{#1}}
\expandafter\ifx\csname urlstyle\endcsname\relax
  \providecommand{\doi}[1]{doi: #1}\else
  \providecommand{\doi}{doi: \begingroup \urlstyle{rm}\Url}\fi

\bibitem[Betker et~al.(2023)Betker, Goh, Jing, Brooks, Wang, Li, Ouyang,
  Zhuang, Lee, Guo, et~al.]{betker2023improving}
James Betker, Gabriel Goh, Li~Jing, Tim Brooks, Jianfeng Wang, Linjie Li, Long
  Ouyang, Juntang Zhuang, Joyce Lee, Yufei Guo, et~al.
\newblock Improving image generation with better captions.
\newblock \emph{Computer Science}, 2023.
\newblock URL \url{https://cdn.openai.com/papers/dall-e-3.pdf}.

\bibitem[Chang et~al.(2022)Chang, Zhang, Barber, Maschinot, Lezama, Jiang,
  Yang, Murphy, Freeman, Rubinstein, et~al.]{chang2022maskgit}
Huiwen Chang, Han Zhang, Jarred Barber, Ajinkya Maschinot, Jose Lezama,
  Lu~Jiang, Ming-Hsuan Yang, Kevin Murphy, William~T Freeman, Michael
  Rubinstein, et~al.
\newblock {MaskGIT}: Masked generative image transformer.
\newblock In \emph{Proceedings of the IEEE/CVF Conference on Computer Vision
  and Pattern Recognition}, pages 11315--11325, 2022.

\bibitem[Chen et~al.(2023)Chen, Yu, Ge, Yao, Xie, Wu, Wang, Kwok, Luo, Lu,
  et~al.]{chen2023pixart}
Junsong Chen, Jincheng Yu, Chongjian Ge, Lewei Yao, Enze Xie, Yue Wu, Zhongdao
  Wang, James Kwok, Ping Luo, Huchuan Lu, et~al.
\newblock {PixArt-$\alpha$}: Fast training of diffusion transformer for
  photorealistic text-to-image synthesis.
\newblock \emph{arXiv preprint arXiv:2310.00426}, 2023.

\bibitem[Chen et~al.(2025)Chen, Wu, Wu, Ma, Liu, Pan, Liu, Xie, Yu, Ruan, and
  Luo]{januspro2025}
Xiaokang Chen, Chengyue Wu, Zhiyu Wu, Yiyang Ma, Xingchao Liu, Zizheng Pan, Wen
  Liu, Zhenda Xie, Xingkai Yu, Chong Ruan, and Ping Luo.
\newblock {Janus-Pro}: Unified multimodal understanding and generation with
  data and model scaling.
\newblock \emph{arXiv preprint arXiv:2501.17811}, 2025.

\bibitem[Chen et~al.(2024{\natexlab{a}})Chen, Qian, Chu, Liu, Zhang, Chen,
  Wang, and Li]{chen2024janus}
Yao Chen, Siyuan Qian, Yuchen Chu, Zheng Liu, Kaipeng Zhang, Qifeng Chen,
  Wenping Wang, and Hao Li.
\newblock Janus: Decoupling visual encoding for unified multimodal
  understanding and generation.
\newblock \emph{arXiv preprint arXiv:2410.13848}, 2024{\natexlab{a}}.

\bibitem[Chen et~al.(2024{\natexlab{b}})Chen, Zhou, Shen, Hong, Sun, Gutfreund,
  and Gan]{hu2024visual}
Zhenfang Chen, Qinhong Zhou, Yikang Shen, Yining Hong, Zhiqing Sun, Dan
  Gutfreund, and Chuang Gan.
\newblock Visual chain-of-thought prompting for knowledge-based visual
  reasoning.
\newblock In \emph{AAAI}, volume~38, pages 1254--1262, 2024{\natexlab{b}}.

\bibitem[Deng et~al.(2025)Deng, Zhu, Li, Gou, Li, Wang, Zhong, Yu, Nie, Song,
  et~al.]{deng2025emerging}
Chaorui Deng, Deyao Zhu, Kunchang Li, Chenhui Gou, Feng Li, Zeyu Wang, Shu
  Zhong, Weihao Yu, Xiaonan Nie, Ziang Song, et~al.
\newblock Emerging properties in unified multimodal pretraining.
\newblock \emph{arXiv preprint arXiv:2505.14683}, 2025.

\bibitem[Esser et~al.(2021)Esser, Rombach, and Ommer]{esser2021taming}
Patrick Esser, Robin Rombach, and Bjorn Ommer.
\newblock Taming transformers for high-resolution image synthesis.
\newblock In \emph{Proceedings of the IEEE/CVF Conference on Computer Vision
  and Pattern Recognition}, pages 12873--12883, 2021.

\bibitem[Esser et~al.(2024)Esser, Kulal, Blattmann, Entezari, M{\"u}ller,
  Saini, Levi, Lorenz, Sauer, Boesel, et~al.]{SD3}
Patrick Esser, Sumith Kulal, Andreas Blattmann, Rahim Entezari, Jonas
  M{\"u}ller, Harry Saini, Yam Levi, Dominik Lorenz, Axel Sauer, Frederic
  Boesel, et~al.
\newblock Scaling rectified flow transformers for high-resolution image
  synthesis.
\newblock In \emph{ICML}, pages 12606--12633, 2024.

\bibitem[Feng et~al.(2023)Feng, Zhu, Fu, Jampani, Akula, He, Basu, Wang, and
  Wang]{wu2024layoutgpt}
Weixi Feng, Wanrong Zhu, Tsu-jui Fu, Varun Jampani, Arjun Akula, Xuehai He,
  Sugato Basu, Xin~Eric Wang, and William~Yang Wang.
\newblock {LayoutGPT}: Compositional visual planning and generation with large
  language models.
\newblock \emph{Advances in Neural Information Processing Systems}, 36, 2023.

\bibitem[Ge et~al.(2023)Ge, Zhao, Zeng, Ge, Li, Wang, and Shan]{ge2023planting}
Yuying Ge, Sijie Zhao, Ziyun Zeng, Yixiao Ge, Chen Li, Xintao Wang, and Ying
  Shan.
\newblock Planting a {SEED} of vision in large language model.
\newblock \emph{arXiv preprint arXiv:2307.08041}, 2023.

\bibitem[Ge et~al.(2024)Ge, Zhao, Zhu, Ge, Yi, Song, Li, Ding, and
  Shan]{seed-x}
Yuying Ge, Sijie Zhao, Jinguo Zhu, Yixiao Ge, Kun Yi, Lin Song, Chen Li,
  Xiaohan Ding, and Ying Shan.
\newblock {SEED-X}: Multimodal models with unified multi-granularity
  comprehension and generation.
\newblock \emph{arXiv preprint arxiv:2404.14396}, 2024.

\bibitem[Ghosh et~al.(2023)Ghosh, Hajishirzi, and Schmidt]{ghosh2023geneval}
Dhruba Ghosh, Hannaneh Hajishirzi, and Ludwig Schmidt.
\newblock {GenEval}: An object-focused framework for evaluating text-to-image
  alignment.
\newblock \emph{Advances in Neural Information Processing Systems},
  36:\penalty0 52132--52152, 2023.

\bibitem[Jin et~al.(2023)Jin, Xu, Chen, , et~al.]{jin2023unified}
Yang Jin, Zhibin Xu, Ying Chen, , et~al.
\newblock Unified language-vision pretraining in {LLM} with dynamic discrete
  visual tokenization.
\newblock \emph{arXiv preprint arXiv:2309.04669}, 2023.

\bibitem[Labs(2024)]{flux}
Black~Forest Labs.
\newblock Flux, 2024.
\newblock URL \url{https://github.com/black-forest-labs/flux}.

\bibitem[Liao et~al.(2025)Liao, Liu, Wang, Luo, Zhang, Zhao, Wu, Li, Tian, and
  Huang]{liao2025mogao}
Chao Liao, Liyang Liu, Xun Wang, Zhengxiong Luo, Xinyu Zhang, Wenliang Zhao,
  Jie Wu, Liang Li, Zhi Tian, and Weilin Huang.
\newblock Mogao: An omni foundation model for interleaved multi-modal
  generation.
\newblock \emph{arXiv preprint arXiv:2505.05472}, 2025.

\bibitem[Liu et~al.(2024)Liu, Yan, Zaharia, and Abbeel]{lwm}
Hao Liu, Wilson Yan, Matei Zaharia, and Pieter Abbeel.
\newblock World model on million-length video and language with ringattention.
\newblock \emph{arXiv preprint arxiv:2402.08268}, 2024.

\bibitem[Lu et~al.(2023)Lu, Clark, Zellers, Mottaghi, and
  Kembhavi]{lu2022unifiedio}
Jiasen Lu, Christopher Clark, Rowan Zellers, Roozbeh Mottaghi, and Aniruddha
  Kembhavi.
\newblock {Unified-IO}: A unified model for vision, language, and multi-modal
  tasks.
\newblock In \emph{International Conference on Learning Representations}, 2023.

\bibitem[Oord et~al.(2017)Oord, Vinyals, and Kavukcuoglu]{oord2017neural}
Aaron van~den Oord, Oriol Vinyals, and Koray Kavukcuoglu.
\newblock Neural discrete representation learning.
\newblock \emph{Advances in Neural Information Processing Systems}, 30, 2017.

\bibitem[OpenAI(2023)]{dalle3}
OpenAI.
\newblock Dall·e 3 system card, 2023.
\newblock \url{https://cdn.openai.com/papers/DALL_E_3_System_Card.pdf}.

\bibitem[{OpenAI}(2024)]{openai2024gpt4o}
{OpenAI}.
\newblock Gpt-4o system card, 2024.
\newblock URL \url{https://openai.com/index/gpt-4o-system-card/}.

\bibitem[Pan et~al.(2025)Pan, Shukla, Singh, Zhao, Mishra, Wang, Xu, Chen, Li,
  Juefei-Xu, et~al.]{pan2025transfer}
Xichen Pan, Satya~Narayan Shukla, Aashu Singh, Zhuokai Zhao, Shlok~Kumar
  Mishra, Jialiang Wang, Zhiyang Xu, Jiuhai Chen, Kunpeng Li, Felix Juefei-Xu,
  et~al.
\newblock Transfer between modalities with {MetaQueries}.
\newblock \emph{arXiv preprint arXiv:2504.06256}, 2025.

\bibitem[Podell et~al.(2024)Podell, English, Lacey, Blattmann, Dockhorn,
  M{\"u}ller, Penna, and Rombach]{sdxl}
Dustin Podell, Zion English, Kyle Lacey, Andreas Blattmann, Tim Dockhorn, Jonas
  M{\"u}ller, Joe Penna, and Robin Rombach.
\newblock {SDXL}: Improving latent diffusion models for high-resolution image
  synthesis.
\newblock In \emph{ICLR}, 2024.

\bibitem[Qu et~al.(2024)Qu, Zhang, Liu, Wang, Jiang, Gao, Ye, Du, Yuan, and
  Wu]{qu2024tokenflow}
Liao Qu, Huichao Zhang, Yiheng Liu, Xu~Wang, Yi~Jiang, Yiming Gao, Hu~Ye,
  Daniel~K Du, Zehuan Yuan, and Xinglong Wu.
\newblock {TokenFlow}: Unified image tokenizer for multimodal understanding and
  generation.
\newblock \emph{arXiv preprint arXiv:2412.03069}, 2024.

\bibitem[Ramesh et~al.(2022)Ramesh, Dhariwal, Nichol, Chu, and Chen]{dalle2}
Aditya Ramesh, Prafulla Dhariwal, Alex Nichol, Casey Chu, and Mark Chen.
\newblock Hierarchical text-conditional image generation with clip latents.
\newblock \emph{arXiv preprint arxiv:2204.06125}, 2022.

\bibitem[Rombach et~al.(2022)Rombach, Blattmann, Lorenz, Esser, and
  Ommer]{rombach2022high}
Robin Rombach, Andreas Blattmann, Dominik Lorenz, Patrick Esser, and Bj{\"o}rn
  Ommer.
\newblock High-resolution image synthesis with latent diffusion models.
\newblock In \emph{CVPR}, pages 10684--10695, 2022.

\bibitem[Saharia et~al.(2022)Saharia, Chan, Saxena, Li, Whang, Denton,
  Ghasemipour, Ayan, Mahdavi, Lopes, et~al.]{saharia2022photorealistic}
Chitwan Saharia, William Chan, Saurabh Saxena, Lala Li, Jay Whang, Emily
  Denton, Seyed Kamyar~Seyed Ghasemipour, Burcu~Karagol Ayan, Seyedeh~Sara
  Mahdavi, Rapha~Gontijo Lopes, et~al.
\newblock Photorealistic text-to-image diffusion models with deep language
  understanding.
\newblock \emph{Advances in Neural Information Processing Systems},
  35:\penalty0 36479--36494, 2022.

\bibitem[Sun et~al.(2023)Sun, Ge, Lu, Chen, Wang, Li, Chen, Zhang, Zhang, Li,
  et~al.]{sun2023emu}
Quanhan Sun, Yuying Ge, Xuyang Lu, Haodong Chen, Peng Wang, Xufeng Li, Kaipeng
  Chen, Jingyi Zhang, Rui Zhang, Aojun Li, et~al.
\newblock Generative pretraining in multimodality.
\newblock \emph{arXiv preprint arXiv:2312.13286}, 2023.

\bibitem[Team(2024)]{chameleon}
Chameleon Team.
\newblock Chameleon: Mixed-modal early-fusion foundation models.
\newblock \emph{arXiv preprint arXiv:2405.09818}, 2024.

\bibitem[Wang et~al.(2024{\natexlab{a}})Wang, Lu, Yang, Huang, Han, Hou, Zhang,
  and Xu]{wang2024illume}
Chunwei Wang, Guansong Lu, Junwei Yang, Runhui Huang, Jianhua Han, Lu~Hou, Wei
  Zhang, and Hang Xu.
\newblock {ILLUME}: Illuminating your llms to see, draw, and self-enhance.
\newblock \emph{arXiv preprint arXiv:2412.06673}, 2024{\natexlab{a}}.

\bibitem[Wang et~al.(2024{\natexlab{b}})Wang, Zhang, Luo, Sun, Cui, Wang,
  Zhang, Wang, Li, Yu, et~al.]{emu3}
Xinlong Wang, Xiaosong Zhang, Zhengxiong Luo, Quan Sun, Yufeng Cui, Jinsheng
  Wang, Fan Zhang, Yueze Wang, Zhen Li, Qiying Yu, et~al.
\newblock {EMU-3}: Next-token prediction is all you need.
\newblock \emph{arXiv preprint arxiv:2409.18869}, 2024{\natexlab{b}}.

\bibitem[Xie et~al.(2024)Xie, Mao, Bai, Zhang, Wang, Lin, Gu, Chen, Yang, and
  Shou]{show-o}
Jinheng Xie, Weijia Mao, Zechen Bai, David~Junhao Zhang, Weihao Wang,
  Kevin~Qinghong Lin, Yuchao Gu, Zhijie Chen, Zhenheng Yang, and Mike~Zheng
  Shou.
\newblock {Show-O}: One single transformer to unify multimodal understanding
  and generation.
\newblock \emph{arXiv preprint arxiv:2408.12528}, 2024.

\bibitem[Xue et~al.(2024)Xue, Shu, Awadalla, Wang, Yan, Purushwalkam, Zhou,
  Prabhu, Dai, Ryoo, et~al.]{li2024blip3}
Le~Xue, Manli Shu, Anas Awadalla, Jun Wang, An~Yan, Senthil Purushwalkam,
  Honglu Zhou, Viraj Prabhu, Yutong Dai, Michael~S Ryoo, et~al.
\newblock {xGEN-MM (BLIP-3)}: A family of open large multimodal models.
\newblock \emph{arXiv preprint arXiv:2408.08872}, 2024.

\bibitem[Yu et~al.(2024)Yu, Cheng, Wang, Kumar, Macherey, Huang, Ross, Esser,
  Bisk, Yang, et~al.]{yu2024language}
Lijun Yu, Yong Cheng, Zhiruo Wang, Vivek Kumar, Wolfgang Macherey, Yanping
  Huang, David~A Ross, Patrick Esser, Yonatan Bisk, Ming-Hsuan Yang, et~al.
\newblock Language model beats diffusion -- tokenizer is key to visual
  generation.
\newblock \emph{arXiv preprint arXiv:2310.05737}, 2024.

\bibitem[Zhang et~al.(2024)Zhang, Zhang, Li, Zhao, Karypis, and
  Smola]{zhang2024multimodal}
Zhuosheng Zhang, Aston Zhang, Mu~Li, Hai Zhao, George Karypis, and Alex Smola.
\newblock Multimodal chain-of-thought reasoning in language models.
\newblock \emph{Transactions on Machine Learning Research}, 2024.

\bibitem[Zhou et~al.(2024)Zhou, Yu, Babu, Tirumala, Yasunaga, Shamis, Kahn, Ma,
  Zettlemoyer, and Levy]{zhou2024transfusion}
Chunting Zhou, Lili Yu, Arun Babu, Kushal Tirumala, Michihiro Yasunaga, Leonid
  Shamis, Jacob Kahn, Xuezhe Ma, Luke Zettlemoyer, and Omer Levy.
\newblock Transfusion: Predict the next token and diffuse images with one
  multi-modal model.
\newblock \emph{arXiv preprint arXiv:2408.11039}, 2024.

\end{thebibliography}

\clearpage

\beginappendix
\FloatBarrier

\section{Training Configurations}

% \begin{table}[t]
% \centering
% \small
% \caption{Training compute budget measured in H100-hours (GPU-hours).}
% \label{tab:compute_budget}
% \begin{tabular}{l S[table-format=6.0] S[table-format=3.1]}
% \toprule
% \textbf{Stage} & {\textbf{H100-hours}} & {\textbf{Share (\%)}} \\
% \midrule
% Pretraining (tokenizer \& Unified model) & 360000 & 75.0 \\
% Continued training (CT)  & 50000  & 10.4 \\
% Supervised fine-tuning (SFT) & 64000 & 13.3 \\
% Reinforcement learning (RLHF) & 6000  & 1.3 \\
% \midrule
% \textbf{Total} & \textbf{480000} & \textbf{100.0} \\
% \bottomrule
% \end{tabular}
% \end{table}

\begin{table}[h!]
\centering
\small
\caption{Training compute budget measured in GPU-hours on high-performance GPUs.}
\label{tab:compute_budget}
\begin{tabular}{l S[table-format=6.0] S[table-format=3.1]}
\toprule
\textbf{Stage} & {\textbf{GPU-hours}} & {\textbf{Share (\%)}} \\
\midrule
Pretraining (tokenizer and unified model) & 360000 & 75.0 \\
Continued training (CT)  & 50000  & 10.4 \\
Supervised fine-tuning (SFT) & 64000 & 13.3 \\
Reinforcement learning (RLHF) & 6000  & 1.3 \\
\midrule
\textbf{Total} & \textbf{480000} & \textbf{100.0} \\
\bottomrule
\end{tabular}
\end{table}

\section{Limitations and Future Work}
While VoT effectively bridges multimodal understanding and generation, its MoT architecture uses a frozen VLM backbone with parallel experts, limiting the system to unidirectional reasoning-to-generation transfer. This design preserves the VLM's existing capabilities, but future work will explore bidirectional optimization strategies that jointly improve understanding and generation without catastrophic forgetting.

\section{Additional Text-to-Image Generation Results}
\begin{figure}[h!]
    \centering
    \includegraphics[width=1.0\linewidth]{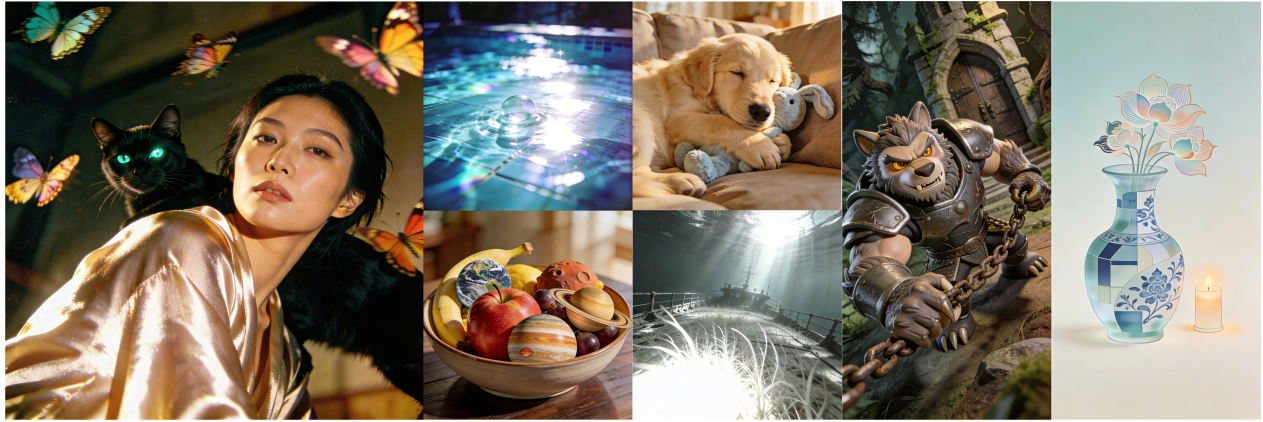}
    \caption{Additional text-to-image generation results.}
    \label{fig:add_t2i}
\end{figure}

\end{document}